\documentclass[runningheads]{llncs}
\usepackage[T1]{fontenc}
\usepackage{graphicx}
\usepackage{xcolor}
\usepackage{amsmath}
\usepackage{amssymb}
\usepackage{array}
\usepackage{multirow}
\usepackage{color}
\usepackage{colortbl}
\usepackage{framed}
\usepackage{bm}
\usepackage{bbm}
\usepackage{xspace}
\usepackage{enumitem}
\usepackage{algorithm}
\usepackage{listings}
\usepackage{url}
\usepackage{booktabs}
\usepackage{pifont} 
\usepackage{lipsum}
\usepackage{bm}
\usepackage{circledsteps}
\usepackage[most]{tcolorbox}
\usepackage{adjustbox}
\usepackage{xcolor}
\usepackage{bbm}
\usepackage{mathtools}
\usepackage{lipsum}
\usepackage{adjustbox}
\usepackage{stfloats}
\usepackage{listings}
\usepackage{hyperref}

\newcommand{\cmark}{\ding{51}}%

\newcommand{\tit}[1]{\smallskip\noindent\textbf{#1.}}
\newcommand{\tinytit}[1]{\noindent\textbf{#1.}}
\newcommand{\ours}{RaTA-Tool\xspace}

\definecolor{LightCyan}{rgb}{0.88,0.95,1}
\definecolor{blond}{HTML}{DBBDD5}
\colorlet{blond}{blond!60}
\definecolor{green}{rgb}{0.0, 0.62, 0.42}
\definecolor{LightGray}{gray}{0.93}
\definecolor{ourcolor}{rgb}{0.925, 0.902, 0.969}
\definecolor{ourcolorl}{rgb}{0.95, 0.945, 0.975}
\definecolor{pastelblue}{rgb}{0.85, 0.95, 0.98}

\def \ie {\emph{i.e.}}

\usepackage{orcidlink}

\newtcolorbox{promptbox1}[1][]{
  enhanced,
  breakable,
  colback=pastelblue!20,
  colframe=pastelblue!20!black,
  leftrule=1.5mm,
  arc=1mm,
  boxrule=0.6pt,
  top=2mm, bottom=2mm, left=3mm, right=3mm,
  fonttitle=\bfseries,
  title=Prompt,
  fontupper=\ttfamily,
  #1,
  before upper={\color{black}},
}

\definecolor{promptboxbg}{HTML}{F2F2F2}

\lstdefinelanguage{LLMPrompt}{
    basicstyle=\ttfamily\scriptsize,
    breaklines=true,
    backgroundcolor={},    
    frame=none,            
    frameround=ffff,
    breaklines=true,
    breakindent=0pt,        
    breakautoindent=true, 
    xleftmargin=2pt,       
    xrightmargin=2pt,      
    aboveskip=3pt,
    belowskip=3pt,
    literate=
        {→}{$\rightarrow$}1
        {—}{---}1
        {“}{``}1
        {”}{''}1
        {…}{...}1
        {ä}{{\"a}}1
}

\newtcblisting{prompt}[1][]{
    enhanced,
    listing only,
    listing options={
        language=LLMPrompt,
        basicstyle=\ttfamily\scriptsize,
        xleftmargin=2pt,
        xrightmargin=2pt,
    },
    colback=promptboxbg,
    colframe=promptboxbg!20!black,
    leftrule=1.5mm,
    arc=1mm,
    boxrule=1pt,
    top=1mm, bottom=1mm, left=.9mm, right=.9mm,
    fonttitle=\bfseries,
    title=#1,
    titlerule=0pt,               
}

\begin{document}
\title{\ours: Retrieval-based Tool Selection with Multimodal Large Language Models}
\titlerunning{\ours: Retrieval-Augmented Tool Selection with MLLMs}
%
\author{Gabriele Mattioli$^{1}$ \and
Evelyn Turri$^{1}$\orcidlink{0009-0005-5668-0839}\and
Sara Sarto$^{1}$\orcidlink{0000-0003-1057-3374}\and
Lorenzo Baraldi$^{2}$\orcidlink{0009-0000-4658-8928} \and\\
Marcella Cornia$^{1}$\orcidlink{0000-0001-9640-9385} \and
Lorenzo Baraldi$^{1}$\orcidlink{0000-0001-5125-4957} \and
Rita Cucchiara$^{1}$\orcidlink{0000-0002-2239-283X}
}
\authorrunning{G. Mattioli et al.}
%
\institute{University of Modena and Reggio Emilia, Italy\\
\email{\{name.surname\}@unimore.it}
\and
University of Pisa, Italy\\
\email{\{name.surname\}@phd.unipi.it}
\vspace{-0.3cm}
}
\maketitle              
\sloppy
\begin{abstract}
Tool learning with foundation models aims to endow AI systems with the ability to invoke external resources -- such as APIs, computational utilities, and specialized models -- to solve complex tasks beyond the reach of standalone language generation. While recent advances in Large Language Models (LLMs) and Multimodal Large Language Models (MLLMs) have expanded their reasoning and perception capabilities, existing tool-use methods are predominantly limited to text-only inputs and closed-world settings. Consequently, they struggle to interpret multimodal user instructions and cannot generalize to tools unseen during training. In this work, we introduce \ours, a novel framework for \emph{open-world multimodal tool selection}. Rather than learning direct mappings from user queries to fixed tool identifiers, our approach enables an MLLM to convert a multimodal query into a structured task description and subsequently retrieve the most appropriate tool by matching this representation against semantically rich, machine-readable tool descriptions. This retrieval-based formulation naturally supports extensibility to new tools without retraining. To further improve alignment between task descriptions and tool selection, we incorporate a preference-based optimization stage using Direct Preference Optimization (DPO). To support research in this setting, we also introduce the first dataset for open-world multimodal tool use, featuring standardized tool descriptions derived from Hugging Face model cards. Extensive experiments demonstrate that our approach significantly improves tool-selection performance, particularly in open-world, multimodal scenarios.

\end{abstract}
\setcounter{footnote}{0}
\section{Introduction}
\label{sec:intro}

Humans possess a remarkable ability to invent, adapt, and employ tools, allowing them to solve problems beyond their innate physical capabilities. As foundation models continue to evolve, AI systems are starting to acquire comparable capabilities. This emerging direction, commonly referred to as \emph{tool learning with foundation models}~\cite{qin2024tool}, seeks to merge the general reasoning abilities of large models with the specialization of external tools. By granting models access to dynamic knowledge bases and computational utilities~\cite{nakano2021webgpt,thoppilan2022lamda}, tool-augmented systems can perform tasks beyond the scope of standalone language generation.

Exploiting the already extended capabilities of Large Language Models (LLMs)~\cite{achiam2023gpt,brown2020language,grattafiori2024llama,yang2025qwen3}, some works integrate in these models tool-use mechanisms to support real-world tool use at scale. This extension has opened the door to a broad range of applications. Within this broader landscape, Multimodal Large Language Models (MLLMs)~\cite{caffagni2024revolution,cocchi2025llava,liu2024improved} play a pivotal role. By jointly processing text, images, audio, and other modalities~\cite{cheng2024videollama,fu2024vita,han2023onellmframeworkalignmodalities,Qwen2.5-Omni}, MLLMs unify diverse tasks under a single architecture and have achieved impressive performance across conversational, analytical, and reasoning applications, particularly when trained with instruction-following paradigms. 

Despite these advances, most existing tool-use frameworks operate under two restrictive assumptions. First, they largely rely on textual inputs~\cite{qin2024toolllm}, limiting their ability to interpret multimodal instructions where visual or audio information is essential for disambiguation. Second, they adopt a closed-world formulation: tool selection is treated as a supervised classification problem over a fixed set of tools observed during training. As a result, these systems struggle to generalize to newly introduced tools or to tasks whose required capabilities were not explicitly represented in the training distribution.
Recent works have started to explore multimodal tool selection by training MLLMs to map multimodal inputs directly to tool identifiers~\cite{wang2025mllmtoolmultimodallargelanguage}. While effective in controlled evaluation settings, such approaches fundamentally inherit the limitations of closed-set learning. In contrast, real-world tool ecosystems are dynamic, continually evolving as new tools are introduced and existing ones are updated. This gap highlights the need for an \emph{open-world} formulation of multimodal tool selection, where models can reason over previously unseen tools without retraining.

To address this challenge, we introduce \ours, a novel retrieval-based framework for open-world multimodal tool selection. Rather than learning direct mappings from user queries to tool identifiers, our approach enables an MLLM to reinterpret a multimodal user query into a structured task description that captures the underlying intent and required capabilities. This task representation is then matched against a collection of semantically rich, machine-readable tool descriptions using embedding-based retrieval, allowing the system to naturally generalize to unseen tools. To further enhance performance and refine generated task descriptions, we incorporate a preference-based optimization stage using Direct Preference Optimization (DPO)~\cite{rafailov2023direct}.

Finally, to support systematic evaluation in this setting, we introduce the first dataset for open-world multimodal tool selection. It features standardized, structured tool descriptions derived from Hugging Face model cards, enabling reproducible and extensible evaluation beyond standard classification pipelines.

\tinytit{Contributions} In summary, our contributions are as follows:
\begin{itemize}[noitemsep, topsep=0pt]
    \item We introduce a novel framework for open-world multimodal tool selection that reasons over structured tool descriptions instead of relying on closed-set, supervised query-tool mappings.
    \item We propose a retrieval pipeline in which an MLLM converts user queries into task descriptions that are matched to tools via embedding-based similarity.
    \item We enhance tool-selection performance through a {preference-based alignment stage} using DPO.
    \item We construct the {first benchmark} tailored for open-world multimodal tool selection, featuring standardized, machine-readable tool descriptions.
\end{itemize}

\section{Related Work}
\label{sec:related}

\tinytit{Multimodal Large Language Models}
Recent years have witnessed rapid progress in extending LLMs beyond text to support multimodal inputs such as images, audio, and video. Early approaches align pretrained language backbones with modality-specific encoders through cross-modal projection and instruction tuning. Representative models such as LLaVA~\cite{NEURIPS2023_6dcf277e,liu2024improved}, BLIP~\cite{pmlr-v202-li23q}, and Flamingo~\cite{10.5555/3600270.3601993} demonstrate strong performance in multimodal perception, reasoning, and generation while largely preserving frozen or lightly adapted encoders.

Subsequent works have extended MLLMs beyond images to additional modalities such as audio and video by aligning pretrained language backbones with modality-specific encoders through adapter or connector modules~\cite{chen2023xllmbootstrappingadvancedlarge,zhao2023chatbridgebridgingmodalitieslarge,pipoli2025missrag}. Complementary efforts focus on video-centric modeling, addressing temporal dynamics and audio-visual fusion via specialized encoders and spatiotemporal alignment mechanisms~\cite{cheng2024videollama,zhang-etal-2023-video}.  More recent approaches move toward unified multimodal perception and generation within a single architecture, supporting tightly coupled reasoning across text, vision, audio, and speech~\cite{fu2024vita,Qwen2.5-Omni}. In parallel, alternative designs explore modality-agnostic representations by removing modality-specific components altogether and introducing universal encoders capable of handling diverse input types~\cite{han2023onellmframeworkalignmodalities}.
While these models significantly enhance multimodal understanding, they primarily focus on improving perception and reasoning within the model itself. They do not explicitly address how multimodal inputs can be leveraged to select and invoke external tools, nor do they consider open-world settings where tools are unseen during training.

\tit{Tool Use with Language Models}
Parallel to advances in LLM capabilities, a growing body of work explores the use of language models as agents that invoke external tools to solve complex tasks. Toolformer~\cite{schick2023toolformer} introduces the idea of teaching LLMs to decide when and how to call APIs through self-supervised annotation. Subsequent systems extend this paradigm to broader application domains such as code generation, web search, healthcare, and task automation. Several works treat tool usage as either retrieve an API or planning actions from a predefined set of tools. Gorilla~\cite{NEURIPS2024_e4c61f57} focuses on accurate API selection in machine learning workflows, while ToolLLM~\cite{qin2024toolllm} emphasizes multi-step task execution using predefined tool inventories. On a different line, systems such as HuggingGPT~\cite{shen2023hugginggpt}, Visual GPT~\cite{wu2023visualchatgpttalkingdrawing} and GPT-4Tools~\cite{yang2023gpttools} decompose user requests into subtasks and route them to specialized models. Despite their effectiveness, these methods primarily operate on textual inputs and therefore cannot directly leverage multimodal cues present in many real-world user instructions.

\tit{Multimodal Tool Agents}
More recently, multimodal tool agents extend tool-use paradigms to MLLMs, enabling direct perception of visual or audio inputs when selecting tools. MLLM-Tool~\cite{wang2025mllmtoolmultimodallargelanguage} represents a first step in this direction by training MLLMs to select tools based on multimodal queries. However, existing approaches largely rely on direct mappings from queries to predefined tool identifiers, inheriting the limitations of closed-set learning and restricting generalization to unseen tools. In contrast, our work addresses \emph{open-world} multimodal tool selection, where relevant tools may not be observed during training. Rather than predicting tool names or IDs, we transform multimodal user queries into structured task descriptions and retrieve suitable tools via semantic matching against rich, machine-readable tool representations. This retrieval-based formulation decouples tool selection from fixed inventories and enables scalable generalization to newly introduced tools.

\begin{figure}[t]
    \centering
    \includegraphics[width=0.98\linewidth]{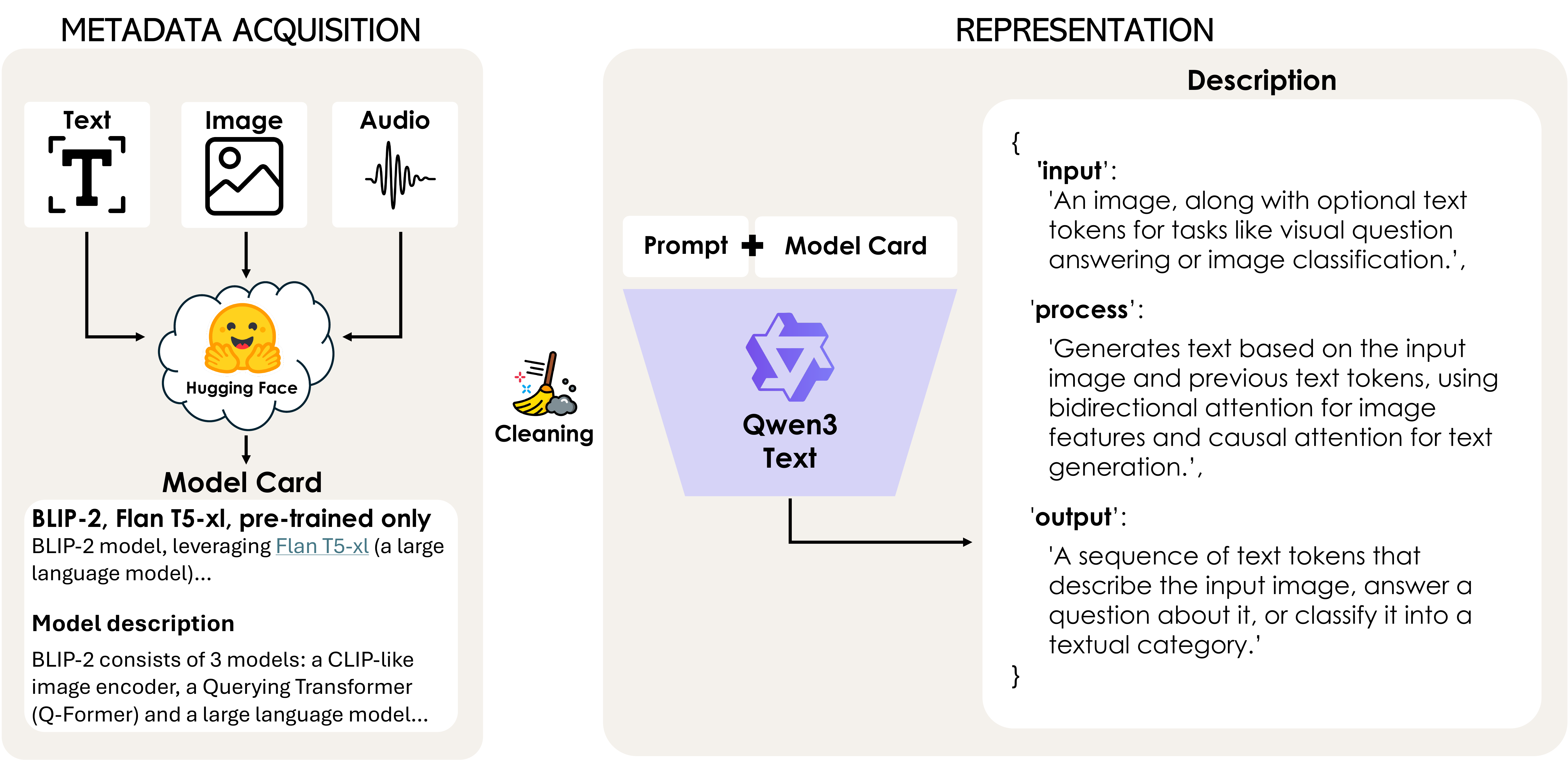}
    \vspace{-0.15cm}
    \caption{Pipeline of dataset creation. We first perform a metadata acquisition step by scraping model cards from Hugging Face. The collected data then undergoes a cleaning stage.
    Finally, given a user query and its associated prompt, we feed this information into an LLM to generate a structured tool description in a standardized JSON format.}
    \label{fig:dataset_pipeline}
    \vspace{-0.2cm}
\end{figure}

\section{Proposed Method}
\label{sec:method}

\subsection{Dataset Creation}
\label{sec:dataset_creation}

\tinytit{Preliminaries} As an initial step, we build upon ToolMMBench~\cite{wang2025mllmtoolmultimodallargelanguage}, a publicly available dataset that links multimodal user queries to Hugging Face models. The dataset is designed to evaluate LLM ability to select external tools when faced with text-ambiguous queries. 

Let $\mathcal{Q} = \{q_1, q_2, \dots, q_M\}$ denote the set of user queries and $\mathcal{T} = \{t_1, t_2, \dots, t_N\}$ the set of candidate tools. ToolMMBench provides an initial collection of query-tool associations organized into modality-specific splits, including \textit{text-only}, \textit{text-image}, and \textit{text-audio} inputs. In each split, a user textual query $q \in \mathcal{Q}$, optionally accompanied by multimodal input, is associated with a single external tool $t \in \mathcal{T}$.

Although this supervised formulation enables controlled evaluation of tool selection, it inherently restricts generalization to tools observed during training. In particular, models trained under this setting can only select from a fixed, closed set of tools, limiting their ability to handle user queries that require previously unseen tools or newly introduced capabilities. This closed-world assumption motivates the need for methods that generalize beyond the training tool set by reasoning over tool descriptions rather than relying on query-tool associations.

Therefore, we use this resource solely as an initial foundation, upon which we perform extensive pre-processing and adaptation to construct a dataset aligned with our experimental setting. To align the original dataset with our experimental setting, we first standardize the structure of all query-tool pairs to match our required format. We then perform dataset-wide cleaning procedures to remove duplicated or inconsistent samples, ensuring that each query is associated with a single, well-defined tool description.

\begin{figure}[t]
    \centering
    \includegraphics[width=0.98\linewidth]{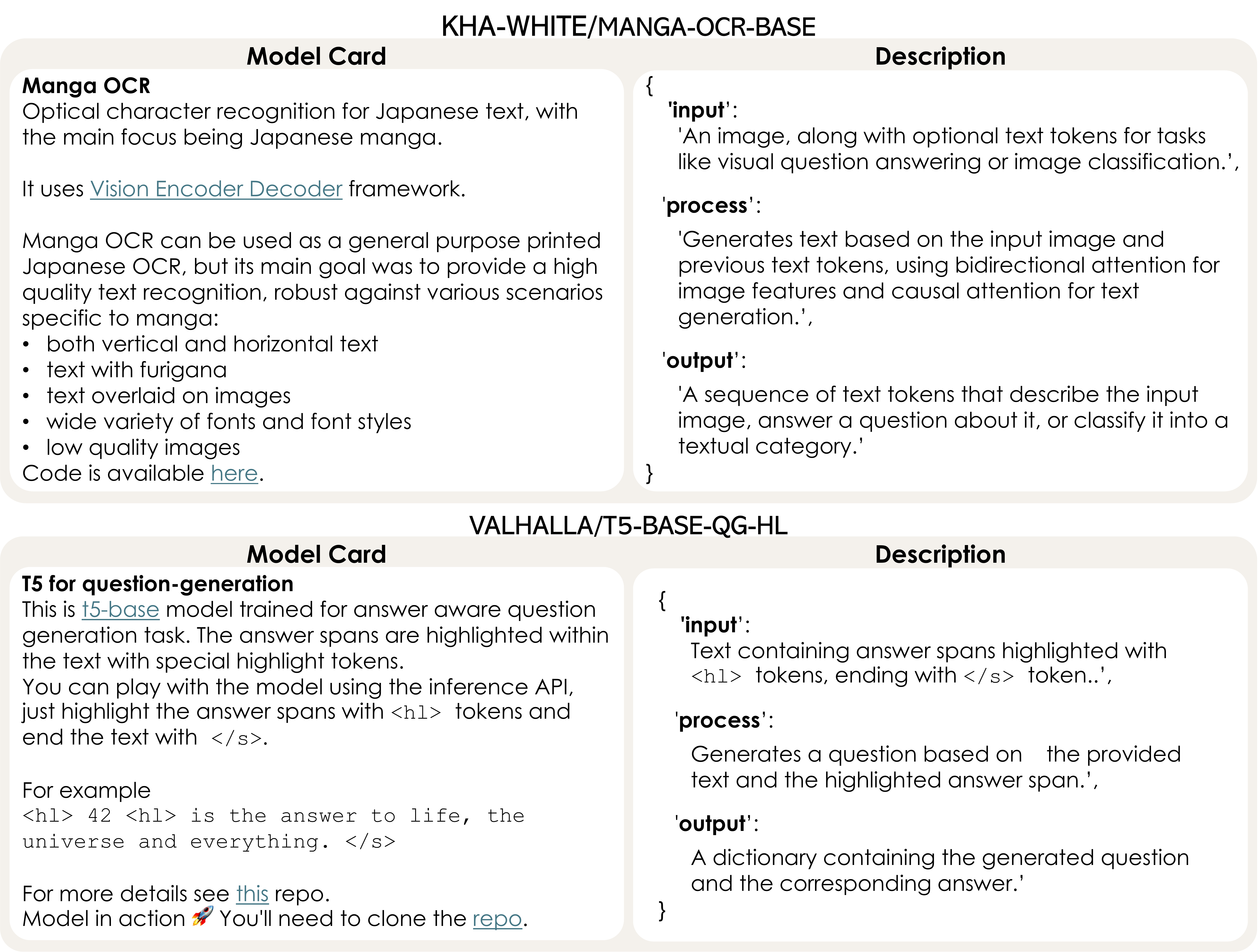}
    \vspace{-0.15cm}
    \caption{Qualitative dataset examples illustrating Hugging Face model cards and the corresponding tool descriptions represented in the standardized JSON format.}
    \label{fig:qualitative_dataset}
    \vspace{-0.2cm}
\end{figure}

\tit{Tool Metadata Acquisition and Representation}
Each tool in our dataset corresponds to a Hugging Face model. To obtain descriptive metadata, we programmatically scraped the model card associated with each tool. Model cards typically contain semi-structured natural language descriptions of the model functionality, intended use cases, input modalities, and expected outputs. However, this information is not directly suitable for automated reasoning or structured tool selection.

To address this limitation, we employ an LLM (\ie, Qwen3~\cite{yang2025qwen3}) to transform each model card into a standardized JSON representation. Specifically, given the user query and the corresponding model card as input, the LLM is prompted to extract and generate a structured description comprising three fields: \Circled{1} \texttt{input}, the expected input format and modalities supported by the tool; \Circled{2} \texttt{output}, the type and structure of the generated output; and \Circled{3} \texttt{process}, a concise functional summary of the tool capabilities.

This structured representation enables consistent and machine-readable tool descriptions, facilitating downstream tool selection and reasoning tasks. An overview of the dataset creation pipeline is shown in Fig.~\ref{fig:dataset_pipeline}, while some qualitative examples from our dataset are shown in Fig.~\ref{fig:qualitative_dataset}.

\begin{figure}[t]
    \centering
    \includegraphics[width=0.99\linewidth]{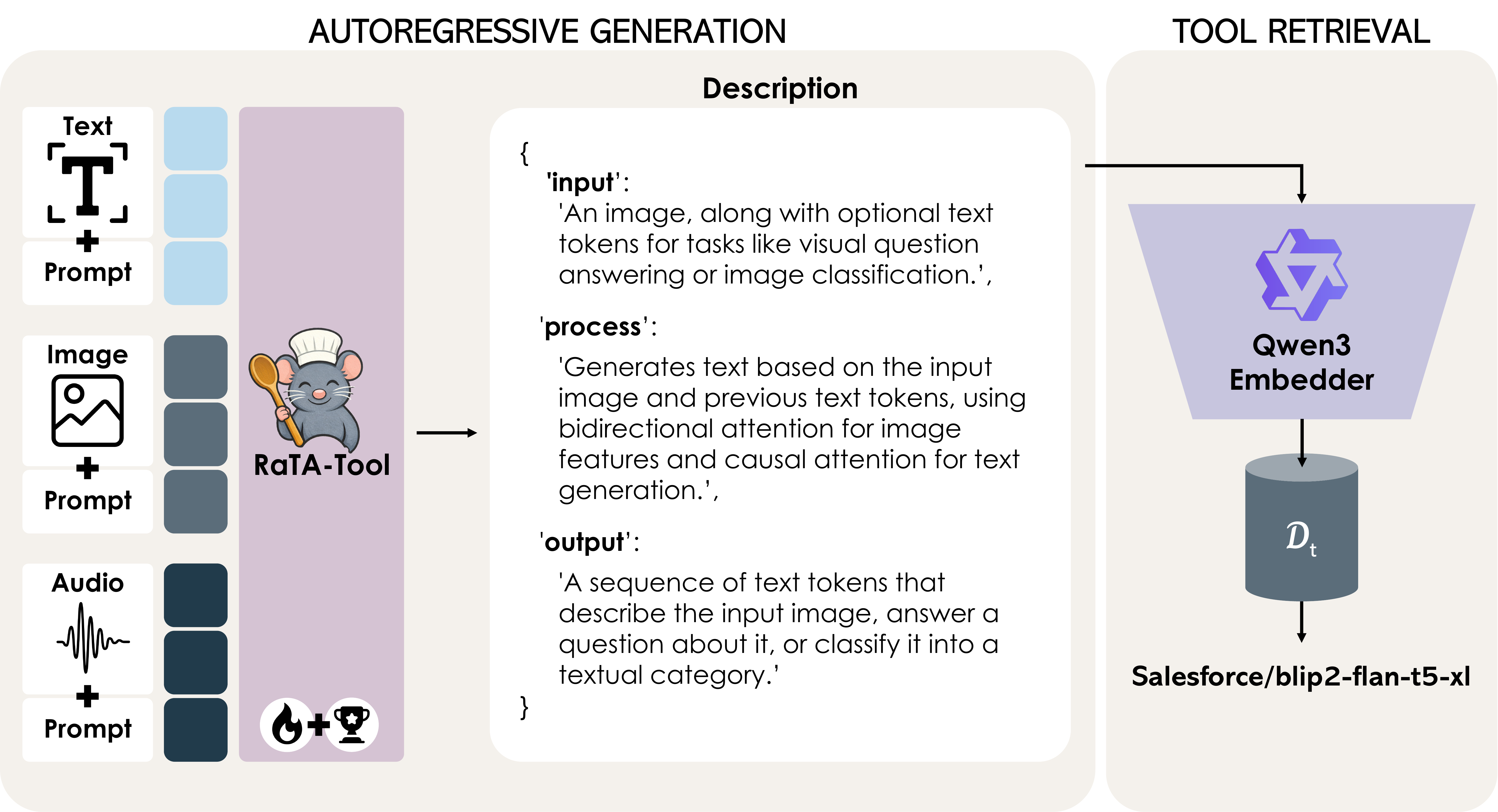}
    \vspace{-0.15cm}
    \caption{Overview of our pipeline. \ours supports multimodal user queries by jointly processing the input prompt and associated modalities. A fine-tuned LLM encodes the combined inputs to generate a structured task description of the user request, further aligned via Direct Preference Optimization (DPO). This description is embedded and used to retrieve the most relevant tool from an external tool collection.}
    \label{fig:method_pipeline}
    \vspace{-0.2cm}
\end{figure}

\subsection{The \ours Framework}

Given an input query $q \in \mathcal{Q}$, our objective is to retrieve the most appropriate tool for a user-requested task, from a collection of candidate descriptions. 
Overall, \ours comprises two components: an MLLM that takes as input the user query and a retrieval module which selects the most relevant tool. An overview of the proposed framework is shown in Fig.~\ref{fig:method_pipeline}.

\tit{Multimodal Integration and Autoregressive Generation}
A MLLM typically receives a query composed of one or more input modalities (for example, text, image, or audio) and generates a textual output in an autoregressive manner. The textual prompt usually includes a pre-defined system-level prompt and a question related to the input modality, given by the user. Specifically for our use case, given the input query and a specific prompt, an MLLM $\mathcal{G}$ is trained via a supervised fine-tuning (SFT) stage to generate a structured task description, following the standardized JSON format of our dataset. Formally, $\tilde{d} = \mathcal{G}(q)$.

Let $\tilde{d} = \{z_1, z_2, \dots, z_T \}$ denote the sequence of tokens composing the task description. Training is performed using a standard autoregressive cross-entropy loss, defined as 
\begin{equation}
    \mathcal{L}_{\text{SFT}} = - \sum_{t=1}^{T} \log p_{\mathcal{G}}(z_t \mid q, z_{<t}),
\end{equation}
where \(p_{\mathcal{G}}(z_t \mid q, z_{<t})\) denotes the conditional probability of generating the next token given the input query and the previously generated tokens.

The generated descriptions capture the expected input, output, and functional intent of the user’s task, forming the basis for subsequent tool retrieval.

\tit{Tool Collection Retrieval}
The external tool collection comprises a set of tool description $\mathcal{D}_t = \{d_{t_1},d_{t_2}, \dots, d_{t_N} \}$, each represented in the standardized JSON format. The goal of the retrieval module is to identify the tool whose description best aligns with the model-generated task description $\tilde{d}$.

To compare descriptions, we embed both  $\tilde{d}$ and each ${d}_t \in \mathcal{D}_t$ using a pretrained embedding model (\ie, Qwen3-Embedding~\cite{qwen3embedding}). Let $E(\cdot)$ denote the embedding function. The similarity between the generated query description and a candidate tool description is modeled as the inner product between their respective embeddings:
\begin{equation}
\label{eq:similarity_computation}
    \text{sim}(\tilde{d}, d_t) = E(\tilde{d}) \cdot E(d_t).
\end{equation}
The selected tool is the one maximizing this similarity score:
\begin{equation}
   t^{*} = \arg\max_{t \in \mathcal{T}} \, \text{sim}(\tilde{d}, d_t).
\end{equation}

\subsection{Tool Selection via Direct Preference Optimization}\label{sec:dpo}

\tinytit{Preliminaries}
Direct Preference Optimization (DPO)~\cite{rafailov2023direct} has emerged as an effective alternative to Reinforcement Learning from Human Feedback (RLHF)~\cite{ouyang2022training} for aligning language models with preference signals. Unlike RLHF, which relies on explicit reward modeling and policy optimization, DPO directly optimizes the model by contrasting preferred and dispreferred responses. Although initially proposed for text-only LLMs, DPO and its variants~\cite{song2024preference,wu2024beta} have been successfully extended to multimodal models~\cite{compagnoni2025mitigating,jia2025symdpo,poppi2026countervid} and a range of generative tasks, including diffusion models~\cite{wallace2024diffusion}, image captioning~\cite{moratelli2024revisiting,sarto2025positive}, and visual question answering~\cite{compagnoni2025reag}. 

In DPO, the policy model is trained to assign higher likelihood to a preferred response $y_w$ than to a less-preferred response $y_l$, relative to a fixed reference model. In our setting, the policy corresponds to the task-description generator $\mathcal{G}$, and the input corresponds to the multimodal user query $q$. The DPO objective is defined as:
\begin{equation}
\label{eq:dpo}
\begin{aligned}
\mathcal{L}_{\text{DPO}}(\pi_\theta; \pi_{\text{ref}}) 
= - \mathbb{E}_{(q, y_w, y_l) \sim \mathcal{D}} \Big[
\log \sigma \Big(
&\beta \log \frac{\pi_\theta(y_w \mid q)}{\pi_{\text{ref}}(y_w \mid q)} \\
&- \beta \log \frac{\pi_\theta(y_l \mid q)}{\pi_{\text{ref}}(y_l \mid q)}
\Big) \Big],
\end{aligned}
\end{equation}
where $\beta$ controls the strength of the regularization that ties the policy $\pi_\theta$ to the frozen reference model $\pi_{\text{ref}}$. In practice, $\pi_\theta$ is initialized from $\pi_{\text{ref}}$.

\tit{Preference Dataset Construction}
To apply DPO in our framework, we construct a preference dataset derived from the SFT-trained model. Given a multimodal query $q$, we generate multiple candidate task descriptions using the SFT model under diverse decoding strategies, including greedy decoding, beam search, and temperature-based sampling. Let $\mathcal{S} = \{s_1, \dots, s_K\}$ denote the resulting set of candidate descriptions.

Let $\mathcal{T} = \{t_1, t_2, \dots, t_N\}$ be the tool collection, and let \(t^* \in \mathcal{T}\) denote the ground-truth tool associated with query $q$. Each candidate description $s_k$ is embedded and compared against all tool descriptions using the similarity function defined in Eq.~\ref{eq:similarity_computation}. Tools are ranked in descending order of similarity, where a lower rank indicates better alignment with the ground-truth tool.

For each query, the candidate description yielding the best (lowest) rank for $t^*$ is selected as the preferred response $y_w$. A non-preferred response \(y_l\) is sampled uniformly from the remaining candidates. Queries for which all candidates achieve identical ranks are discarded to avoid ambiguous preference signals. The resulting preference pairs are then used to fine-tune the model via the DPO objective defined in Eq.~\ref{eq:dpo}.

\section{Experiments}

\subsection{Experimental Setting}

\tinytit{Dataset Details}
For the SFT stage, we build upon ToolMMBench~\cite{wang2025mllmtoolmultimodallargelanguage} and apply the pre-processing and standardization steps described in Sec.~\ref{sec:dataset_creation}. To generate the ground-truth descriptions in JSON format starting from the Hugging Face model cards, we use Qwen3-8B\footnote{\href{https://huggingface.co/Qwen/Qwen3-8B}{\texttt{Qwen/Qwen3-8B}}}~\cite{yang2025qwen3}. The resulting dataset is partitioned into training and test splits. Dataset statistics are reported in Table~\ref{tab:data_statistic}.

The split is performed at the \emph{tool level} to reflect the open-world evaluation setting. Specifically, 90\% of the tools are assigned to the training split and the remaining 10\% to the test split, yielding 826 training tools and 94 test tools, respectively. Queries are assigned to a split based on the associated tool, ensuring that tools appearing in the test set are never observed during training. Particular care is taken to preserve the distribution of input modalities (text-only, text-image, and text-audio) across splits.

For DPO training, we construct a preference dataset comprising 2,981 training items and 409 evaluation items, following the procedure described in Sec.~\ref{sec:dpo}. For each query, we generate five candidate responses using diverse decoding strategies to encourage output variability.  Specifically, we employ greedy decoding, beam search with a beam width of 5, and three sampling-based configurations: medium-temperature sampling (using temperature equal to 0.7), high-temperature sampling (with temperature equal to 1.0), and sampling combined with beam search using three beams. This mix of deterministic and stochastic decoding strategies yields a diverse set of candidate responses from which preference pairs are constructed.

\tit{Training and Evaluation Protocols} Each dataset item is defined as a pair consisting of a user query and a tool. A query may be purely textual or accompanied by an additional modality, such as an image or audio input. As a consequence, multiple items may share the same textual prompt while differing in their associated multimodal inputs, and are therefore treated as distinct queries. Moreover, the dataset naturally exhibits a many-to-many relationship: a single tool can be associated with multiple queries, and conversely, a single query may correspond to different tools depending on the intended task.

During training, this multiplicity is preserved, allowing the model to observe multiple valid tool associations for similar or identical queries. At test time, each query-tool pair is evaluated independently and exactly once. The model predicts a single tool for each query, and performance is measured by whether the retrieved tool matches the ground-truth tool associated with that query instance. Detailed statistics on the number of queries and tools in each split are provided in Table~\ref{tab:data_statistic}. Overall, the dataset consists of 14,924 unique queries.

\begin{table}[t]
  \centering
    \caption{Dataset statistics.}
  \label{tab:data_statistic}
  \vspace{-0.1cm}
  \setlength{\tabcolsep}{.4em}
  \resizebox{0.9\textwidth}{!}{
  \begin{tabular}{lc cccc c cccc}
    \toprule
    & &  \multicolumn{4}{c}{\textbf{Unique Queries}} && \multicolumn{4}{c}{\textbf{Unique Tools}}\\
    \cmidrule{3-6} \cmidrule{8-11}
    & &  Text & Image & Audio & All & & Text & Image & Audio & All \\
    \midrule
     Training split & & 6,381 & 3,575 & 625 & 10,581 & & 585 & 170 & 71  & 826 \\
     Test split & & 3,231 & 922 & 190 & 4,343 & & 66 & 20 & 8 & 94 \\
     \midrule
     Overall & & 9,612 & 4,497 & 815 & 14,924 & & 651 & 190 & 79  & 920 \\
    \bottomrule
  \end{tabular}
  }
  \vspace{-0.35cm}
\end{table}

\tit{Implementation and Training Details}
For structured task description generation, we adopt Qwen2.5-Omni models~\cite{Qwen2.5-Omni} with two different scales, namely 3B and 7B parameters\footnote{\href{https://huggingface.co/Qwen/Qwen2.5-Omni-3B}{\texttt{Qwen/Qwen2.5-Omni-3B}} and \href{https://huggingface.co/Qwen/Qwen2.5-Omni-7B}{\texttt{Qwen/Qwen2.5-Omni-7B}}}, as the underlying MLLM. Both variants are fine-tuned using Low-Rank Adaptation (LoRA)~\cite{hu2022lora} with rank $r=8$ and scaling factor $\alpha=16$. Supervised fine-tuning is performed for 4 epochs on 2 NVIDIA L40S/A40 GPUs, using a learning rate of $2\times10^{-4}$ and a weight decay of $0.01$. We further align the models via DPO, again employing LoRA with the same configuration as in SFT. DPO training is conducted for 8 epochs on a single NVIDIA L40S/A40 GPU, with a learning rate of $5\times10^{-5}$ and a weight decay of $0.01$. For the retrieval stage, we use Qwen3-Embedding-8B\footnote{\href{https://huggingface.co/Qwen/Qwen3-Embedding-8B}{\texttt{Qwen/Qwen3-Embedding-8B}}}~\cite{qwen3embedding} as the text embedding model to encode both generated task descriptions and tool descriptions.

\begin{table}[t]
  \centering
    \caption{Accuracy results on aggregate metrics: unweighted average across modalities (\(\text{Avg}_q\)) and weighted average (\(\text{Avg}_m\)). 
    }
      \vspace{-0.1cm}
  \label{tab:main_results}
  \setlength{\tabcolsep}{.25em}
  \resizebox{\textwidth}{!}{
  \begin{tabular}{lc ccc cc ccc c cc}
    \toprule
    & & & & \multicolumn{2}{c}{\textbf{Training}} & & \multicolumn{3}{c}{\textbf{Modalities}} \\
    \cmidrule{5-6} \cmidrule{8-10}
    & & \textbf{Retriever} & & SFT & DPO & & Text & Image & Audio & & \textbf{\(\text{Avg}_q\)} & \textbf{\(\text{Avg}_m\)} \\
    \midrule
    Qwen2.5-Omni-3B~\cite{Qwen2.5-Omni} & & Contriever & & - & - & & 17.7 & 8.0 & 36.3 & & 16.5 & 20.7 \\
    Qwen2.5-Omni-7B~\cite{Qwen2.5-Omni} & & Contriever & & - & - & & 18.3 & 6.5 & 27.9 & & 16.2 & 17.6 \\
    \midrule
    Qwen2.5-Omni-3B~\cite{Qwen2.5-Omni} & & Qwen3-Embedding & & - & - & & 24.9 & 31.3 & 69.0 & & 28.2 & 41.7 \\
    Qwen2.5-Omni-7B~\cite{Qwen2.5-Omni} & & Qwen3-Embedding & & - & - & & 25.1 & 32.1 & 67.4 & & 28.4 & 41.5 \\
    \midrule
     & & Contriever & & \cmark & - & & 57.1 & 37.0 & 80.5 & & 53.8 & 58.2 \\
     & & Qwen3-Embedding & & \cmark & - & & 71.0 & 55.7 & 82.5 & & 68.3 & 69.7 \\
    \rowcolor{blond}
    \textbf{\ours-3B} & & Qwen3-Embedding & & \cmark & \cmark & & \textbf{71.6} & \textbf{56.1} & \textbf{83.7} & & \textbf{68.8} & \textbf{70.4} \\
    \midrule
     & & Contriever & & \cmark & - & & 51.2 & 31.8 & 79.0 & & 48.3 & 54.0 \\
     & & Qwen3-Embedding & & \cmark & - & & 69.3 & \textbf{58.8} & \textbf{83.7} & & 67.7 & 70.6 \\
    \rowcolor{blond}
    \textbf{\ours-7B} & & Qwen3-Embedding & & \cmark & \cmark & & \textbf{71.6} & 57.1 & \textbf{83.7} && \textbf{69.1} & \textbf{70.8} \\
    \bottomrule
  \end{tabular}
  }
  \vspace{-0.35cm}
\end{table}

\subsection{Experimental Results}
Table~\ref{tab:main_results} reports accuracy results on the test set of the introduced dataset using the standardized JSON tool-description format. Results are shown separately for text, image, and audio queries, together with two aggregate metrics: an unweighted average across modalities (\(\text{Avg}_q\)) and a weighted average (\(\text{Avg}_m\)), where weights reflect the number of samples per modality. This evaluation protocol allows us to assess modality-specific behavior while also capturing overall system performance. We evaluate two embedding backbones for the retrieval stage. Our main configuration relies on Qwen3-Embedding~\cite{qwen3embedding}, while we additionally consider the Contriever model~\cite{izacard2021unsupervised}, based on the BERT base version~\cite{devlin2019bert}, to assess the impact of the embedding model on tool selection.

In the zero-shot setting, performance is generally limited, particularly when using Contriever. With this retrieval model, both Qwen2.5-Omni backbones achieve low aggregate accuracy, with \(\text{Avg}_m\) below 21 for the 3B model and even lower for the 7B variant (\ie, 17.6). The modality breakdown reveals uneven behavior: audio queries reach moderate accuracy, whereas text and image queries remain substantially lower. Replacing Contriever with Qwen3-Embedding yields a large improvement, increasing \(\text{Avg}_m\) to 41.7 (3B) and 41.5 (7B). The gains are consistent across modalities and especially pronounced for image and audio, indicating that retrieval quality and modality coverage of the embedding space are critical even without task-description fine-tuning.

SFT substantially improves performance for both model sizes, confirming the importance of aligning the generator to produce structured task descriptions in the standardized JSON format. Again, the retriever choice remains critical: for the 3B setting, Qwen3-Embedding improves \(\text{Avg}_m\) from 58.2 (Contriever) to 69.7; for 7B, it increases \(\text{Avg}_m\) from 54.0 to 70.6. These differences are reflected across modalities, with particularly gains for text and image queries.

Finally, applying DPO yields further improvements. For \ours-3B, DPO increases performance to \(\text{Avg}_q=68.8\) and \(\text{Avg}_m=70.4\), reaching \(71.6\) (text), \(56.1\) (image), and \(83.7\) (audio). For \ours-7B, DPO produces the best overall results with \(\text{Avg}_q=69.1\) and \(\text{Avg}_m=70.8\), and consistently strong accuracy across modalities. Overall, these results show that (i) a strong embedding model is essential for retrieval-based tool selection, and (ii) preference-based alignment provides complementary gains beyond supervised fine-tuning.

\begin{table}[t]
  \centering 
  \caption{Ablation study on the effect of tool-description format and prompting or fine-tuning strategy on tool-selection performance, reported for both retrieval backbones (Contriever and Qwen3-Embedding). Results are shown for text, image, and audio queries, together with unweighted (\(\text{Avg}_q\)) and weighted (\(\text{Avg}_m\)) aggregate metrics.}
  \label{tab:ablations}
    \vspace{-0.1cm}
  \setlength{\tabcolsep}{.3em}
  \resizebox{\textwidth}{!}{
  \begin{tabular}{lcc cccc c cc c cc}
    \toprule
    & & & \multicolumn{2}{c}{\textbf{Descriptions}} & & \multicolumn{3}{c}{\textbf{Modalities}} \\
    \cmidrule{4-5} \cmidrule{7-9}
    & \textbf{Retriever} & & Type & Mode & & Text & Image & Audio & & \textbf{\(\text{Avg}_q\)} & \textbf{\(\text{Avg}_m\)} \\
    \midrule
    & Contriever & & NL & Zero-shot & & 16.8 & 17.2 & 43.7  && 18.1 & 25.9\\
    & Contriever & & NL & Few-shot & & 13.3 & 13.7 & 49.5 && 15.0 & 25.5 \\
    \cmidrule{2-12}
    & Contriever & & JSON & Zero-shot & & 18.3 & 6.5 & 27.9 && 16.2 & 17.6 \\
    & Contriever & & JSON & Few-shot & & 21.0 & 15.1 & 35.8 && 20.4 & 24.0 \\
    \cmidrule{2-12}
    & Contriever & & NL & SFT & & 34.3 & \textbf{50.3} & 79.0 && 39.6 & \textbf{54.5} \\
    \rowcolor{blond}
     \textbf{\ours-7B} & Contriever & & JSON & SFT & & \textbf{51.2} & 31.8 & \textbf{79.0} & & \textbf{48.3} & 54.0 \\
    \midrule
    & Qwen3-Emb & & NL & Zero-shot & & 27.6 & 32.0 & 72.1 && 30.5 & 43.9 \\
    & Qwen3-Emb & & NL & Few-shot  & & 31.5 & 22.1 & 72.1 && 31.3 & 41.9 \\
    \cmidrule{2-12}
    & Qwen3-Emb & & JSON & Zero-shot & & 25.1 & 32.1 & 67.4 && 28.4 & 41.5 \\
    & Qwen3-Emb & & JSON & Few-shot  & & 25.9 & 38.7 & 70.0 && 30.6 & 44.9 \\
    \cmidrule{2-12}
    & Qwen3-Emb & & NL & SFT & & 41.6 & 44.0 & \textbf{95.8} && 44.5 & 60.5 \\
     \rowcolor{blond}
     \textbf{\ours-7B} & Qwen3-Emb & & JSON & SFT & & \textbf{69.3} & \textbf{58.8} & 83.7 & & \textbf{67.7} & \textbf{70.6} \\
    \bottomrule
  \end{tabular}
  }
  \vspace{-0.35cm}
\end{table}

\subsection{Ablation and Analysis}
Table~\ref{tab:ablations} presents an ablation study using Qwen2.5-Omni-7B as the task-description generator, focusing on the impact of the tool-description format and the prompting or fine-tuning strategy. Results are reported for both retrieval backbones previously considered, namely Contriever and Qwen3-Embedding.

We first analyze the effect of the description format under zero-shot and few-shot prompting. Across both retrievers, structured JSON descriptions outperform natural-language (NL) ones. For example, with Qwen3-Embedding in the zero-shot setting, JSON achieves an \(\text{Avg}_m\) of \(41.5\), compared to \(43.9\) for NL, while under few-shot prompting JSON improves to \(44.9\), narrowing and reversing the gap relative to NL (\(41.9\)). A similar trend is observed with Contriever, where JSON under few-shot prompting (\(\text{Avg}_m=24.0\)) outperforms NL (\(\text{Avg}_m=25.5\)).

Comparing prompting strategies, few-shot prompting consistently improves over zero-shot inference for both NL and JSON formats, with especially notable gains for image and audio queries. For instance, when using Qwen3-Embedding with JSON descriptions, image accuracy increases from \(32.1\) to \(38.7\) when moving from zero-shot to few-shot prompting. These improvements suggest that in-context examples help the model better capture task-relevant constraints and generate more discriminative task descriptions for retrieval. SFT leads to the largest performance improvements across both retrieval backbones. With Contriever, SFT with JSON descriptions improves \(\text{Avg}_m\) from \(24.0\) (few-shot) to \(54.0\), while with Qwen3-Embedding it increases from \(44.9\) to \(70.6\). In both cases, JSON descriptions outperform NL under SFT, confirming the advantage of standardized, machine-readable formats.

Overall, these results indicate that while retrieval quality remains important, the choice of description format and training strategy plays a central role in effective multimodal tool selection. The best results are achieved by combining SFT with structured JSON descriptions, consistently across both retrieval backbones.

\begin{figure}[t]
    \centering
    \includegraphics[width=0.99\linewidth]{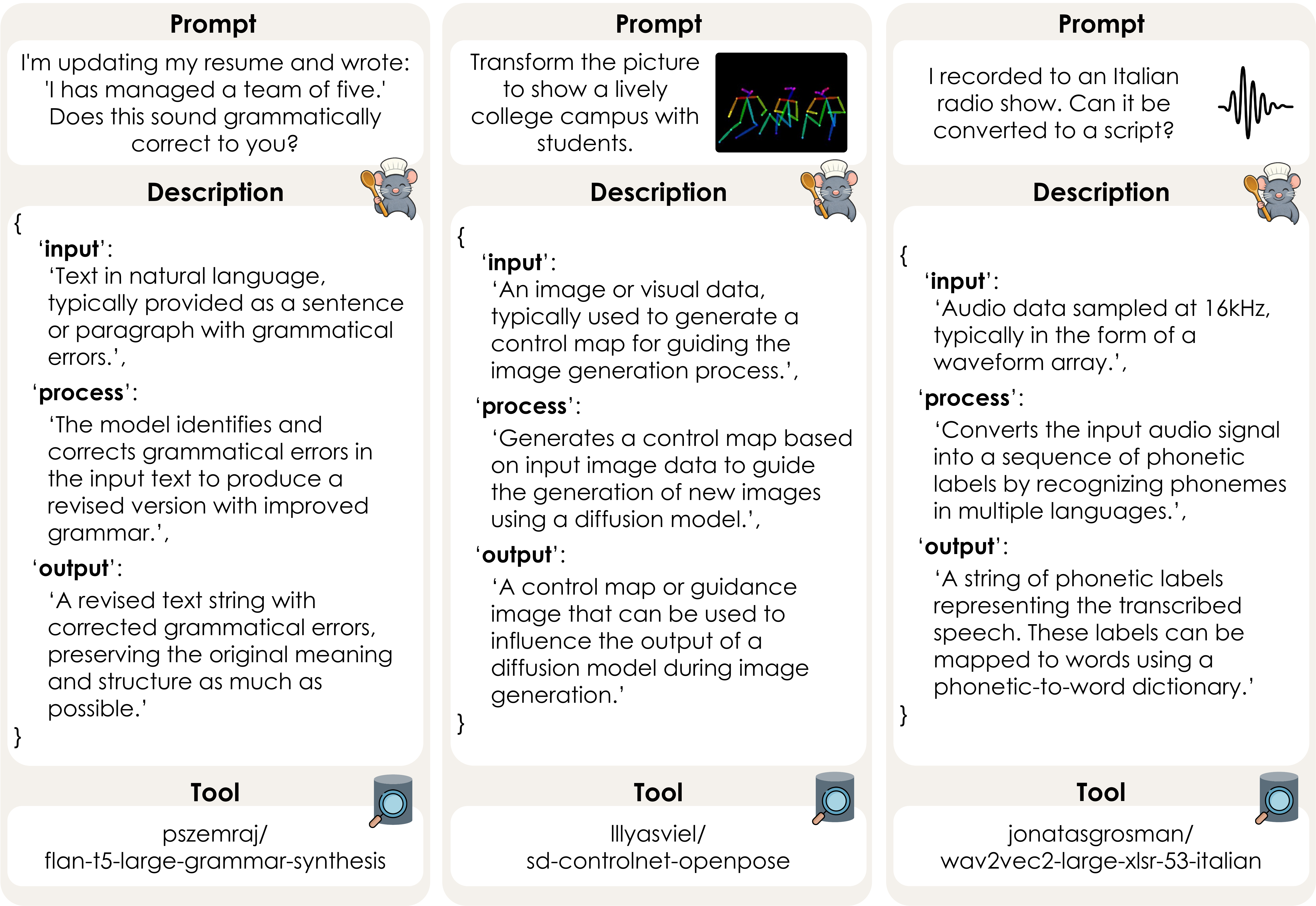}
    \vspace{-0.15cm}
    \caption{Qualitative results of \ours under different input modalities.}
    \vspace{-0.3cm}
    \label{fig:qualitative_model_gen}
\end{figure}

\subsection{Qualitative Results}
We present qualitative results of \ours in Fig.~\ref{fig:qualitative_model_gen}. The examples include purely textual prompts (first column) as well as multimodal prompts (second and third columns). In all cases, the model generates a structured JSON description of the user request, which is then used to retrieve the most relevant tool from the external collection. The generated descriptions accurately capture the expected inputs, functional intent, and outputs of the task, demonstrating effective understanding of both textual and multimodal cues. Across all shown examples, \ours retrieves the correct tool. Additional qualitative examples, together with the exact prompts used in our experiments, are provided in the supplementary material.

\section{Conclusion}
We presented \ours, a retrieval-based framework for open-world multimodal tool selection that reasons over structured, machine-readable tool descriptions instead of relying on closed-set supervision. By converting multimodal user queries into structured task descriptions and retrieving tools via semantic similarity, our approach naturally generalizes to unseen tools. We further showed that preference-based alignment with DPO complements standard SFT, improving robustness and consistency across modalities. To support evaluation in this setting, we introduced the first benchmark for open-world multimodal tool use with standardized tool descriptions. Our results demonstrate that semantic, description-driven tool reasoning is a promising direction for scalable and robust tool-learning systems.



\subsubsection{Acknowledgments} This work has been supported by EU Horizon project ELLIOT (No. 101214398) and by the EuroHPC JU project IT4LIA (No. 101234224). We also acknowledge CINECA for the availability of high-performance computing resources, and for funding Evelyn Turri's PhD.

%
%
%
\bibliographystyle{splncs04}
\bibliography{bibliography}

\clearpage
\appendix
\section{Qualitative Example}
\begin{figure}
    \centering
    \includegraphics[width=0.99\linewidth]{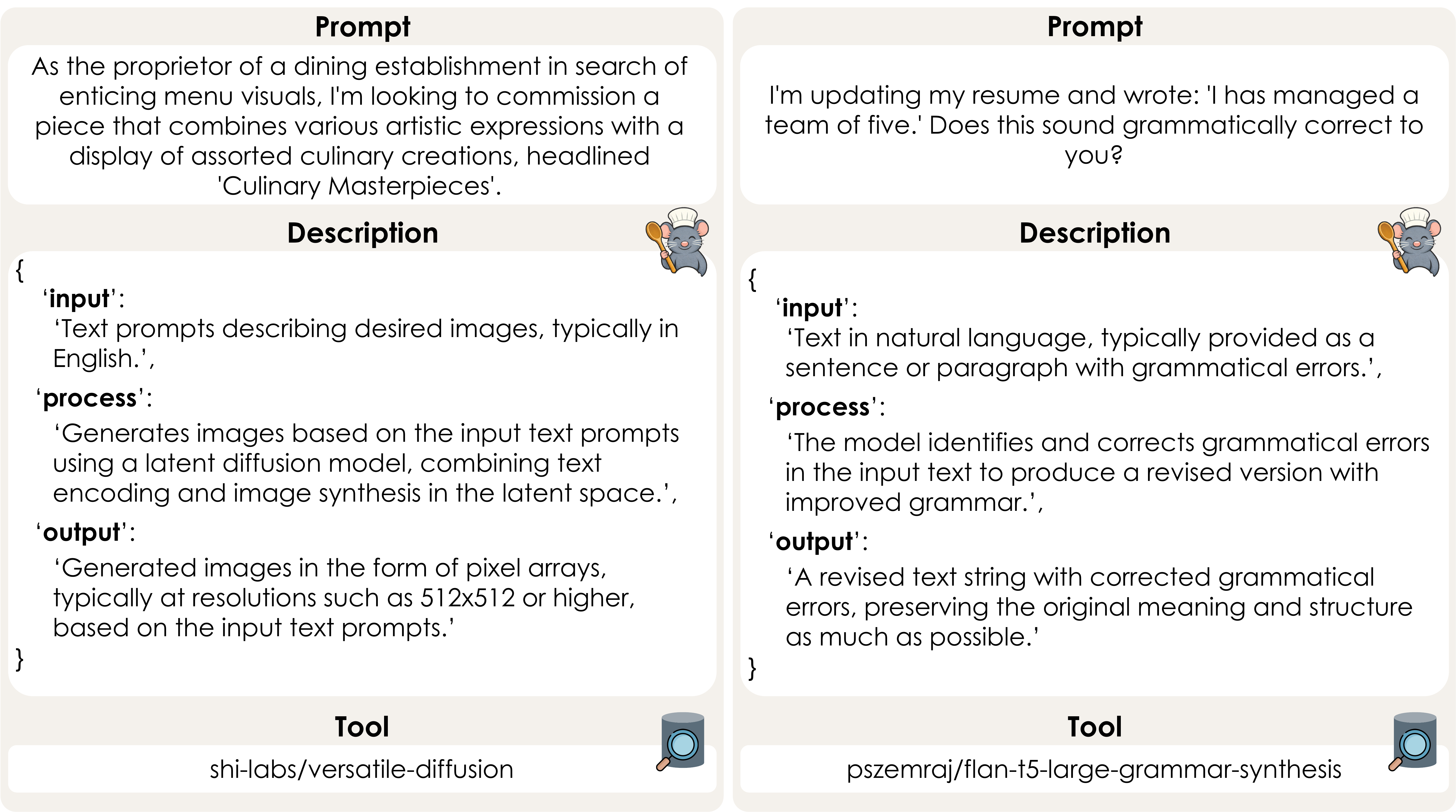}
    \vspace{-0.15cm}
    \caption{Additional qualitative examples of \ours for text-only user queries.}
    \label{fig:qualitative_supplementary_text}
\end{figure}

\begin{figure}
    \centering
    \includegraphics[width=0.99\linewidth]{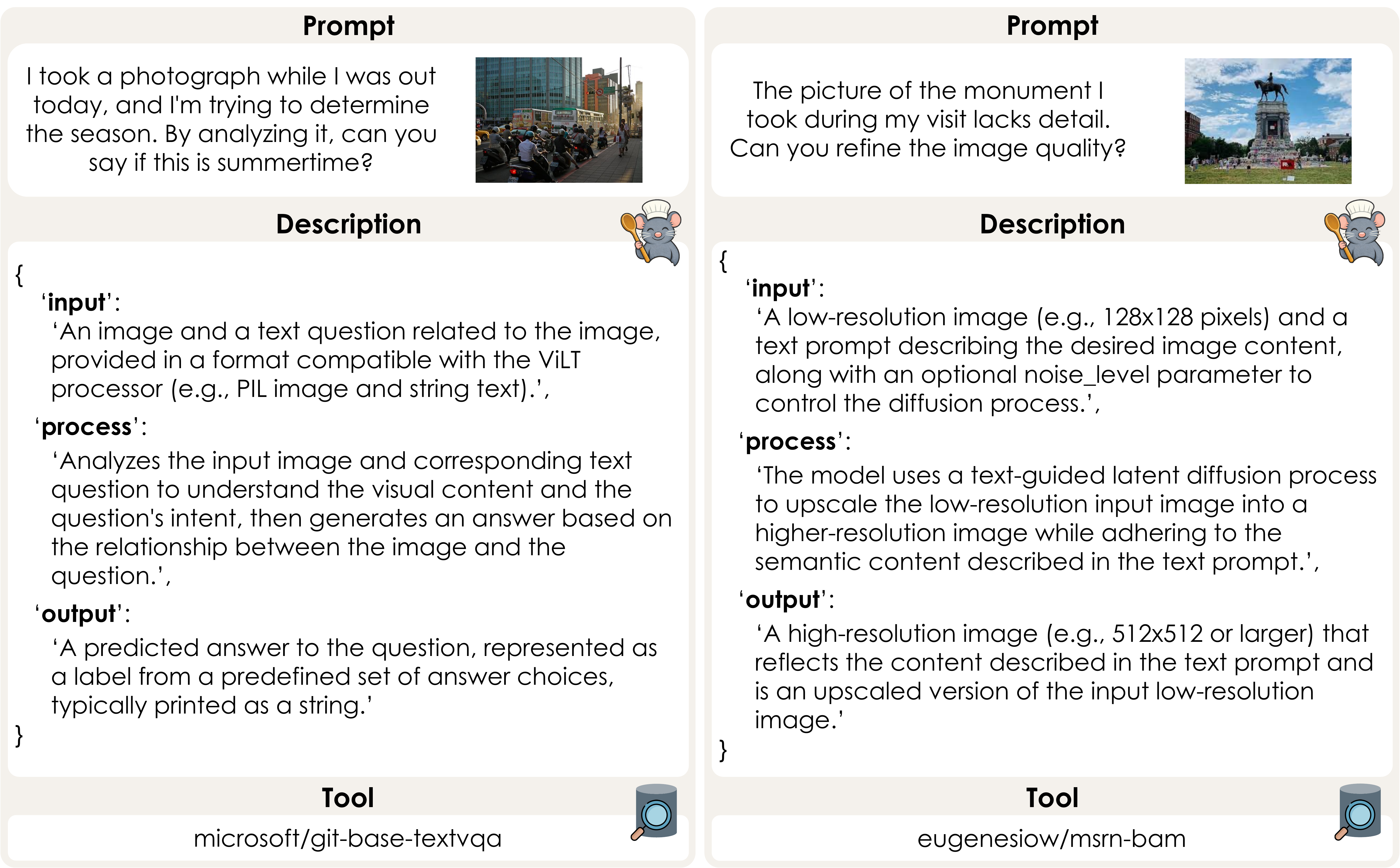}
    \vspace{-0.15cm}
    \caption{Additional qualitative examples of \ours for multimodal user queries with image inputs.}
    \label{fig:qualitative_supplementary_img}
\end{figure}

Additional qualitative examples of \ours are presented in Fig.~\ref{fig:qualitative_supplementary_text} and Fig.~\ref{fig:qualitative_supplementary_img}. These examples include both text-only and multimodal queries. In all cases, \ours generates well-structured, high-quality task descriptions that accurately capture the user intent and lead to the correct tool being retrieved.

\begin{figure}
    \centering
    \includegraphics[width=0.99\linewidth]{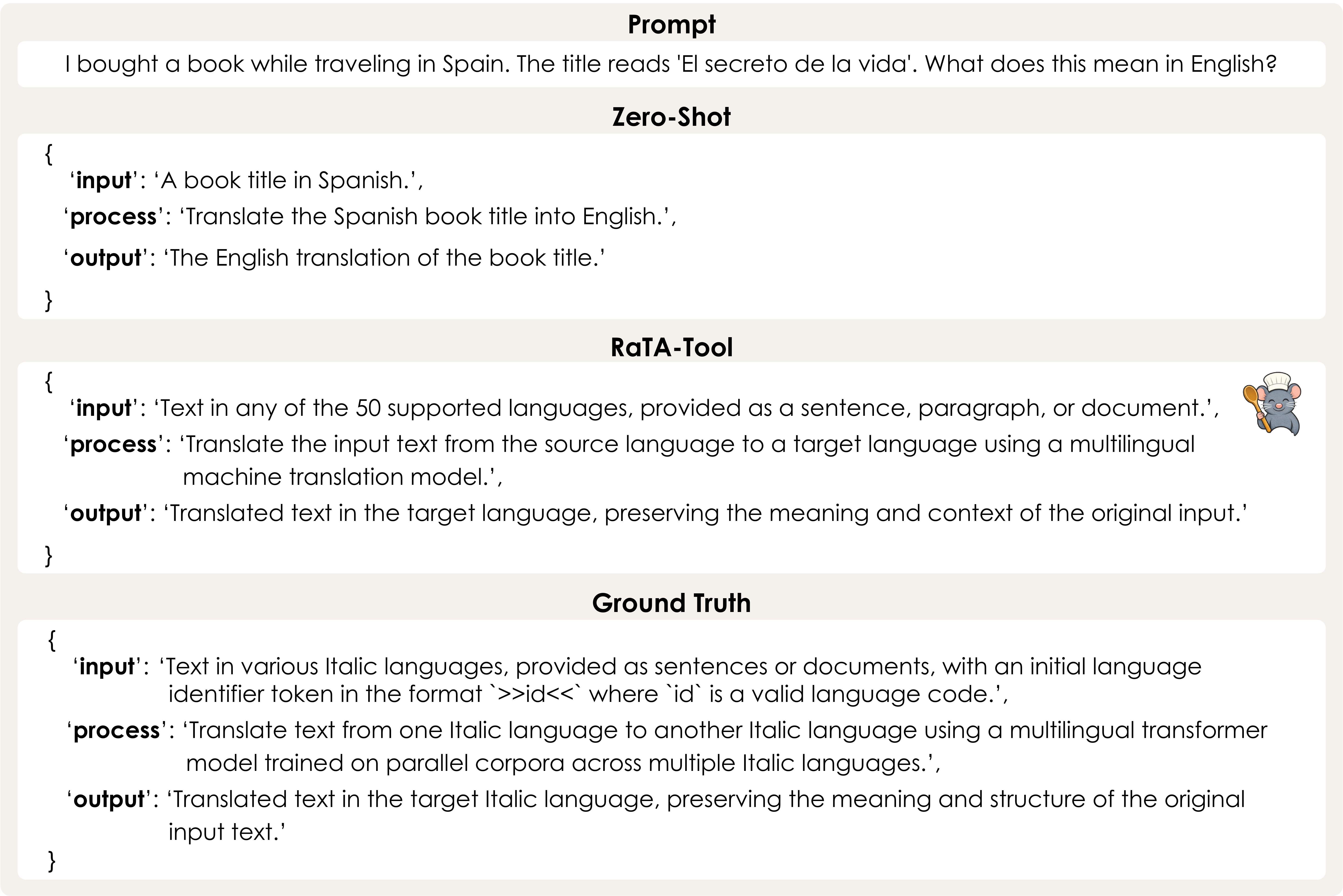}
    \vspace{-0.15cm}
    \caption{Qualitative comparison of task descriptions with zero-shot inference and by \ours, along with the ground-truth descriptions.}
    \label{fig:qualitative_supplementary_zero-shot}
\end{figure}

Fig.~\ref{fig:qualitative_supplementary_zero-shot} further compares task descriptions generated under different training strategies. The results show that \ours not only understands how to structure the description with the desired format, but also produces more detailed and semantically precise descriptions, enabling more reliable retrieval.

\section{Prompts}
We provide examples of the prompts used for task-description generation during dataset creation, considering both structured JSON and natural-language (NL) formats. For dataset construction, the prompt includes a single in-context example to guide the generation of tool descriptions. In contrast, during inference with \ours, the input query is prompted with two in-context examples: one text-only example and one example containing a multimodal input. Fig.~\ref{fig:json_dataset_generation_prompt} and Fig.~\ref{fig:json_input_query} illustrate the JSON-style prompts used for dataset creation and model inference, respectively, while Fig.~\ref{fig:nl_example} shows the corresponding NL-style prompt.

\begin{figure*}
\centering
\begin{prompt}[JSON Model Description Prompt]
You are an AI assistant tasked with extracting structured information from Hugging Face model cards.
Given the model card below, analyze it thoroughly and generate a JSON object that describes the model in terms of:
1. Input: What type of data the model takes in (e.g., text, image, audio, structured data, etc.), including format or shape if mentioned.
2. Process: What the model does with the input (e.g., sentiment classification, translation, object detection, etc.), described concisely.
3. Output: What the model produces as output, including the format, labels or value types where applicable.

Your output must be:
- Model-agnostic (do not mention brand names or model IDs)
- Written strictly in English
- A valid JSON object in the following format:
{
    "input": "<description of the expected input>",
    "process": "<brief description of what the model does>",
    "output": "<description of the output produced>"
}

EXAMPLE
Model card: DeepPavlov/rubert-base-cased HuggingFace model card

Model description:
{
    "input": "Text written in Russian, provided as a sentence, paragraph, or document.",
    "process": "Analyze the linguistic and semantic content of the Russian text and transform it into a multidimensional vector representation that captures meaning, syntax, and contextual nuances.",
    "output": "A numerical embedding vector that encodes the semantic information of the input text, suitable for downstream tasks such as similarity comparison, clustering, or classification."
}

Now do the same for the model card below:
Model card: {model_card}
Model description:
\end{prompt}
\vspace{-0.15cm}
\caption{Example of a JSON-style prompt used for model description generation during dataset creation.}
 \label{fig:json_dataset_generation_prompt}
\vspace{-5pt}
\end{figure*}

\begin{figure*}[t]
\vspace{-5pt}
\centering
\begin{prompt}[JSON Inference Prompt]
Given the user query below, write a complete and precise description of the task requested by the user in JSON format. The JSON object must have three fields:
- input: The nature and format of the input the model will receive.
- process: The process or transformation the model is expected to perform.
- output: The nature and format of the output the model should produce.

Each field must contain short, clear sentences written in English. Do not include specific details about the current query; keep the description applicable to similar tasks of this type.

Your output must be a valid JSON object in the following format:
{
    "input": "<general description of the input>"
    "process": "<general description of the reasoning or steps>"
    "output": "<general description of the expected output>"
}

EXAMPLES

Query 1: I'm watching a WWII movie with a German phrase `Hände hoch. Ich kann dir nicht wehtun.' Can you translate this into Spanish for me?

Task Description 1:
{
    "input": "Text in any of the West Germanic languages, including Afrikaans, Dutch, German, and others, provided as sentences or documents. The input must include a language identifier token in the format `>>id<<' where `id' corresponds to a valid target language.",
    "process": "Translate text from one West Germanic language to another West Germanic language using a multilingual transformer model trained on parallel corpora.",
    "output": "Translated text in the target West Germanic language, corresponding to the input text. The output format is text, and the translation quality is measured using metrics like BLEU and chr-F."
}

Query 2: Show this picture as a group of friends gathered around a campfire, telling stories.

Query 2 Image: {query_2_image}

Task Description 2:
{
    "input": "An image in the form of a normal map, which represents surface orientation information using a 3D vector for each pixel, typically in a grayscale or RGB format.",
    "process": "The model processes the normal map input to guide the diffusion model in generating an image that aligns with the surface orientation information provided by the normal map.",
    "output": "A generated image that adheres to the surface orientation details specified in the normal map input, typically in RGB format."
}

Now do the same for the query below:
Query: {query}
\end{prompt}
\vspace{-2pt}
\caption{Example of a JSON-style prompt used by \ours for task-description generation at inference time.}
 \label{fig:json_input_query}
\vspace{-5pt}
\end{figure*}

\begin{figure*}[t]
\centering
\begin{prompt}[NL Model Description Prompt]
You are an AI assistant tasked with summarizing Hugging Face model cards.
Given the Model Card below, analyze it thoroughly and write a concise, natural language description that explains what kind of input the model expects, what it does with that input, and what it produces as output.

Write the description in clear and complete sentences, using natural language. Avoid technical formatting such as JSON or bullet points. Focus only on information that is explicitly stated or clearly implied in the model card.
The description must be written strictly in English.

EXAMPLE

Model card: 
    DeepPavlov/rubert-base-cased HuggingFace model card

Model description:
    This model processes Russian language text by converting it into a dense semantic vector representation, allowing the meaning of the text to be captured and expressed as an embedding.

Now do the same for the model card below:
Model card: {model_card}
Model description:

\end{prompt}
\begin{prompt}[NL Inference Prompt]
Given the user query below, write a complete and precise description of the task requested by the user. Your description must clearly specify:
- The nature and format of the input the model will receive
- The process or transformation the model is expected to perform
- The nature and format of the output the model should produce
Write the description in clear and direct natural language using full sentences. Do not use code, lists, or technical notation.

Write strictly in English.

EXAMPLES

Query 1: I'm watching a WWII movie with a German phrase `Hände hoch. Ich kann dir nicht wehtun.' Can you translate this into Spanish for me?

Task description 1: This model processes text in various West Germanic languages by translating input text from one West Germanic language into another West Germanic language. The input text must include a language identifier at the beginning of each sentence to specify the source language. The model transforms the input by generating a corresponding translation in the target language. The output is the translated text in the specified target West Germanic language.

Query 2: Show this picture as a group of friends gathered around a campfire, telling stories.

Query 2 image: {query_2_image}

Task description 2: This model processes visual input in the form of normal maps, which are images that encode surface orientation information. It is designed to work in conjunction with a text-to-image diffusion model to guide the generation of images based on the provided normal map. The model interprets the normal map to influence the texture and structure of the generated image, producing a realistic visual output that aligns with the input conditions. The output is a fully rendered image that reflects both the textual description and the spatial orientation details provided by the normal map.

Now do the same for the query below:
Query: {query}
\end{prompt}
\vspace{-0.15cm}
\caption{Examples of NL-style prompts used for model description generation during dataset creation and for task-description generation with \ours at inference time.}
\label{fig:nl_example}
\end{figure*}

\end{document}